\def\year{2022}\relax
\documentclass[letterpaper]{article} 
\usepackage{aaai22}  
\usepackage{times}  
\usepackage{helvet}  
\usepackage{courier}  
\usepackage[hyphens]{url}  
\usepackage{graphicx} 
\usepackage{natbib}  
\usepackage{caption} 
\DeclareCaptionStyle{ruled}{labelfont=normalfont,labelsep=colon,strut=off} 
\usepackage[table,xcdraw]{xcolor}

\title{Interpretable Privacy Preservation of Text Representations Using Vector Steganography}
\author{
    Geetanjali Bihani
}
\affiliations{
    Purdue University\\

    gbihani@purdue.edu
}

\begin{document}

\maketitle

\begin{abstract}
Contextual word representations generated by language models (\textit{LMs}) learn spurious associations present in the training corpora. Recent findings reveal that adversaries can exploit these associations to reverse-engineer the private attributes of entities mentioned within the corpora. These findings have led to efforts towards minimizing the privacy risks of language models. However, existing approaches lack interpretability, compromise on data utility and fail to provide privacy guarantees. Thus, the goal of my doctoral research is to develop interpretable approaches towards privacy preservation of text representations that retain data utility while guaranteeing privacy. To this end, I aim to study and develop methods to incorporate steganographic modifications within the vector geometry to obfuscate underlying spurious associations and preserve the distributional semantic properties learnt during training.

\end{abstract}

\section{Introduction}
\noindent General purpose language models trained on large corpora of natural language text are prone to learning generalizations from the underlying data \cite{coavoux2018privacy}. Recent studies have shown that language variety contains cues that reveal private attributes of individuals, such as identity, political inclination and gender \cite{nguyen_learning_2021}. These cues allow adversaries to exploit the language models and uncover private attributes of users who speak, write or are mentioned within the text used to train the models \cite{coavoux2018privacy}. These findings have led to an increased interest in privacy preservation research for natural language processing. 

\section{Research Problem}
Current privacy preservation methods for NLP rely on the irreversible removal of rich social signals from data, rendering a normalized text representation geometry \cite{nguyen_learning_2021}. These methods can be divided into three groups: (1) Differential privacy based models (2) Cryptographic models and (3) Adversarial Learning models. While adversarial privacy preservation fails to address explicit leakages, differential privacy preservation fails to defend against implicit leakages \cite{huang-etal-2020-texthide}. In theory, provable privacy defense against both implicit and explicit leakage can be provided by cryptographic methods such as multi-party computations (\textit{MPC}) and homomorphic encryption, but due to their computational burden and loss of data utility, these methods have been been deemed inefficient to be applicable within the deep learning paradigm \cite{huang-etal-2020-texthide}. To combine the benefits of cryptography and deep learning, \citet{huang-etal-2020-texthide} recently proposed adding an instance encoding step while fine-tuning language models in a federated learning setting. Although this approach provided better data utility and faster computations than \textit{MPC} and homomorphic encryption \cite{huang-etal-2020-texthide}, it has been shown to succumb to reconstruction attacks \cite{carlini2021private}. These findings suggest that although cryptographic enhancements to deep learning models might provide significant benefits in terms of privacy guarantees, their current adaptations lack reliability, data utility and computational efficiency. Furthermore, while partially meeting privacy objectives, these models lack explainability and interpretability, failing to concurrently meet fairness and trust objectives within ethical NLP \cite{pruksachatkun2021proceedings}.
In the past, privacy preservation in various media has been achieved through information hiding, using various cryptographic techniques \cite{krishnan_overview_2017}. Steganography is one such technique, which hides secret information within a given medium, without altering the structure of the medium itself. To our knowledge, text steganography approaches for privacy preservation have been limited to text character or bit manipulation, because minor modifications in the text form lead to semantic anomalies \cite{krishnan_overview_2017}. For multi dimensional representations, steganalysis methods such as one described in \citet{jiang_novel_2008} use intensity and imperceptibility features to cover and uncover secret messages within the vector space. Since language models map individual word occurrences to multi-dimensional text representations, utilizing steganography in multi-dimensional spaces for privacy preservation in text representations still remains unexplored.

\section{Completed Work}
Our prior work on analyzing text representation geometries showed that developing a general criteria to target spurious associations in text representations is a challenge, largely due to the dependence of distributional text representation spaces on the underlying model choices and training corpora \cite{bihani_modelchoices, bihani_low_2021}. Our study of attributive word associations in text representation models revealed that models trained on the same text corpora using different context aggregation schemes encode different word associations \cite{bihani_modelchoices}. In another recent paper \cite{bihani_low_2021}, we found that general purpose language models encode varying degrees of word-sense associations across model layers. Specifically, bidirectional language models encode stronger associations between the same polysemes of word occurrences as compared to unidirectional models. Moreover these associations are better retained in lower model layers and are significantly reduced in the upper layers due to contextualization of representations. These results indicate that we cannot create an exhaustive list of the word associations that need to be targeted for privacy preservation \emph{a priori}.

\section{Plan of Research}
To address the gaps in prior works and design trustworthy models of privacy preservation in NLP, I aim to utilize steganography within the text representation space to concurrently preserve data utility and privacy. Hence, I propose the following research question for my doctoral research; 
\textit{\textbf{RQ:} How effectively can steganographic models of privacy preservation obfuscate spurious associations (implicit and explicit) without compromising data utility?} My goal is to minimize the normalization of signals learnt from the training data while ensuring privacy on demand. I propose to achieve this by combining steganography and micro-aggregation within the vector space to obfuscate spurious associations in the vector space geometry i.e. Vector space steganography (\textit{VSS}). This approach will allow the retention of the distributional semantics signals learnt from the data, maintaining data utility and accuracy, while hiding the latent confounds within the vector space, allowing privacy preservation. Moreover, \textit{VSS} will allow interpretability, working towards building trustworthy privacy preserving NLP techniques. Towards this goal, I define three research objectives. My first research objective is--\textbf{RO1:} \emph{Designing VSS applicable on-top-of general purpose language model representations to obfuscate spurious associations.} This objective addresses the privacy preservation part of our research question. To obfuscate any spurious associations related to a given word \emph{w}, steganographic candidates will be generated from the vector space geometry by reversing the steganalysis criteria described for multidimensional vectors in \cite{jiang_novel_2008}. These candidates will be micro-aggregated to create an obfuscated representation for \emph{w}, hiding any latent confounds associated with \emph{w}. My second research objective is--\textbf{RO2:} \emph{Evaluating the data utility post-\textit{VSS}  across general purpose language model representations}. This objective addresses the data utility aspect of my research question. To measure data utility, I will compare the proposed approach to the baseline accuracy thresholds for various NLU tasks as described in \cite{huang-etal-2020-texthide}. My final research objective is--\textbf{RO3:} \emph{Evaluating the provable privacy guarantees of VSS.} This objective will address the efficiency of \textit{VSS} in terms of its privacy guarantees, where we will consider the definitions and baselines provided in \cite{carlini2021private}. Through these objectives, I aim to make progress towards developing methods of interpretable privacy assurance for text representation geometries.

\bibliography{aaai22.bib}
\end{document}